\documentclass[10pt,conference]{IEEEtran}

\usepackage{cite}

\ifCLASSINFOpdf
   \usepackage[pdftex]{graphicx}
   \graphicspath{{figs/}}
   \DeclareGraphicsExtensions{.pdf,.jpeg,.png}
\else
   \usepackage[dvips]{graphicx}
   \graphicspath{{../figs/}}
   \DeclareGraphicsExtensions{.eps}
\fi
\usepackage[cmex10]{amsmath}
\usepackage{amsthm}
\usepackage{amssymb}
\newtheorem{definition}{Definition}

\newtheorem{proposition}{Proposition}

\usepackage{algorithmic}

\usepackage{array}

\ifCLASSOPTIONcompsoc
  \usepackage[caption=false,font=normalsize,labelfont=sf,textfont=sf]{subfig}
\else
  \usepackage[caption=false,font=footnotesize]{subfig}
\fi
\usepackage{url}

\newif\iffinal
\finaltrue
\newcommand{\cmtid}{74}

\iffinal
\else
\usepackage[switch]{lineno}
\fi

\begin{document}
%
\title{The Role of Cyclopean-Eye in Stereo Vision}


\iffinal


%
\author{
    \IEEEauthorblockN{
        Sherlon Almeida da Silva\IEEEauthorrefmark{1}\IEEEauthorrefmark{2},
        Davi Geiger\IEEEauthorrefmark{2},
        Luiz Velho\IEEEauthorrefmark{3} and
        Moacir Antonelli Ponti\IEEEauthorrefmark{1} 
    }
    \IEEEauthorblockA{
        \IEEEauthorrefmark{1}Instituto de Ciências Matemáticas e de Computação, Universidade de São Paulo (ICMC-USP), São Carlos, SP, Brazil
    }
    \IEEEauthorblockA{
        \IEEEauthorrefmark{2}Courant Institute of Mathematical Sciences, New York University (NYU), New York, NY, United States
    }
    \IEEEauthorblockA{
        \IEEEauthorrefmark{3}Instituto de Matemática Pura e Aplicada (IMPA), Rio de Janeiro, RJ, Brazil
    }
    \IEEEauthorblockN{
        \textit{\{sherlon.a, dg1\}@nyu.edu}, \textit{lvelho@impa.br}, \textit{moacir@icmc.usp.br}
    }
}

\else
  \author{SIBGRAPI Paper ID: \cmtid \\ }
  \linenumbers
\fi

\maketitle

\begin{abstract}
This work investigates the geometric foundations of modern stereo vision systems, with a focus on how 3D structure and human-inspired perception contribute to accurate depth reconstruction. We revisit the Cyclopean Eye model and propose novel geometric constraints that account for occlusions and depth discontinuities. Our analysis includes the evaluation of stereo feature matching quality derived from deep learning models, as well as the role of attention mechanisms in recovering meaningful 3D surfaces. Through both theoretical insights and empirical studies on real datasets, we demonstrate that combining strong geometric priors with learned features provides internal abstractions for understanding stereo vision systems.
\end{abstract}


\IEEEpeerreviewmaketitle

\section{Introduction}
\label{sec:introduction}
This paper addresses the problem of stereo vision, recovering a 3D scene from the left (L) and right (R) images of the scene. With the advent of deep learning (DL) and the availability of datasets, a high level of accuracy has been achieved in stereo by RAFT-Stereo~\cite{lipson2021raft}, CREStereo~\cite{li2022practical}, DLNR~\cite{zhao2023high}, Selective-IGEV~\cite{wang2024selective}, FoundationStereo~\cite{wen2025foundationstereo},
DEFOM-Stereo~\cite{jiang2025defom}, and MonoStereo~\cite{cheng2025monster}. In these approaches, during training the input is a stereo image pair, and the loss function is based on the discrepancy between the model predicted disparity map and the ground truth (GT) disparity. Once trained, a DL-based stereo model can infer the disparity from a new pair of images. In close examination, these methods first extract features from the left and right images and then perform disparity estimation at each pixel. Despite DL-based stereo having the best performance, there is a lack of understanding how to go from features to disparity output. Understanding a problem means uncovering meaningful internal representations that capture its structure, as is the case of image stereo features. Understanding not only enables better generalization to novel situations, but also facilitates broader application of the underlying principles. Our goal is to deepen our understanding of stereo vision in addition to feature extraction. By developing abstract 3D representations of reality, we aim in the future to improve the performance and efficiency of stereo algorithms and also to lay the groundwork for using such 3D models in other areas of computer vision.



\subsection{Geometrical Ideas in Stereo Vision}

In our pursuit of a deeper understanding of stereo vision, we adopt a Bayesian framework in which the problem is formulated as estimating the most probable 3D surface, $S^{3D}$, given a stereo pair of input images, $I^L$ and  $I^R$.
Specifically, stereo vision aims to solve:
{\small 
\begin{align}
    \arg \max_{S^{3D}} P(S^{\textrm{3D}}|I^L, I^R) =\arg \max_{S^{3D}} \left [ P(I^L, I^R|S^{\textrm 3D})\, P(S^{\textrm3D})\right ]\nonumber 
\end{align}
}
where $P(I^L, I^R|S^{\textrm{3D}})$  models the likelihood of observing the image pair given a hypothesized 3D surface, and  $P(S^{\textrm{3D}})$ is a prior distribution that encapsulates geometric regularities of natural surfaces.

\subsection{Brief History of Geometric Ideas in Stereo}
The study of such geometric principles long predates Bayesian formalism. As early as 1499, Leonardo da Vinci, in Treatise on Painting, observed that different parts of the background become occluded depending on whether one views the scene with the left or right eye~\cite{davinci2014treatise}. Later, Hermann von Helmholtz~\cite{Helmholtz1910} laid foundational work in binocular vision, emphasizing its role in depth perception. He introduced the concept of the cyclopean eye—a notional single eye positioned centrally on the head—representing the brain’s fused interpretation of binocular input. This idea also serves as a useful geometric abstraction: a coordinate system centered between the eyes for representing 3D surfaces.

Further contributions came from Julesz~\cite{Julesz71}, who championed the cyclopean framework and a bottom-up approach to stereo vision, later advanced by Marr and Poggio~\cite{MarrPoggio79} through a computational model of stereopsis. These studies, and much of the computational modeling up to the early 2000's, laid a conceptual foundation that preceded the current dominance of purely data-driven DL methods.



\subsection{Our Contribution}
In this work, we revisit and extend geometric ideas to model $P(S^{\textrm3D})$, present in equation~\eqref{eq:bayesian-stereo-feature}, and analyze these ideas with data from the Middlebury dataset~\cite{scharstein2014high} and RAFT-Stereo~\cite{lipson2021raft} features that yield $FM_*$. 

Broadly presented, our main contribution is:

\noindent-- \textbf{Foundational Contribution to Vision Understanding:} deepens our understanding of 3D scene geometry, paving the way for more interpretable, generalizable and reliable vision systems.

More specifically, the contributions are as follows.

\noindent-- \textbf{Features:} an analysis of the quality of the features learned and extracted by the DL algorithms from the L/R images.

\noindent-- \textbf{Geometric Constraints:} a proposal of novel geometric constraints for stereo vision. 

\noindent-- \textbf{Cyclopean Eye Model:} the role of it in better understanding the geometry of stereo vision. 

\noindent-- \textbf{Attention:} we argue for the need of an attention mechanism to recover 3D surfaces.

\section{The Cyclopean 3D View  }
\label{sec:model}
We begin by briefly reviewing the Cyclopean coordinate system (XD), which provides a natural framework to describe the geometry of the 3D world. In particular, the relation between the match of the left and right features and the depth assignment is elucidated, so that in the next section we propose novel geometrical constraints. 

\subsection{Cyclopean Coordinate System (XD)}
\label{sec:cyclopean}
\begin{figure}[ht]
   \centering
   \includegraphics[width=\columnwidth]{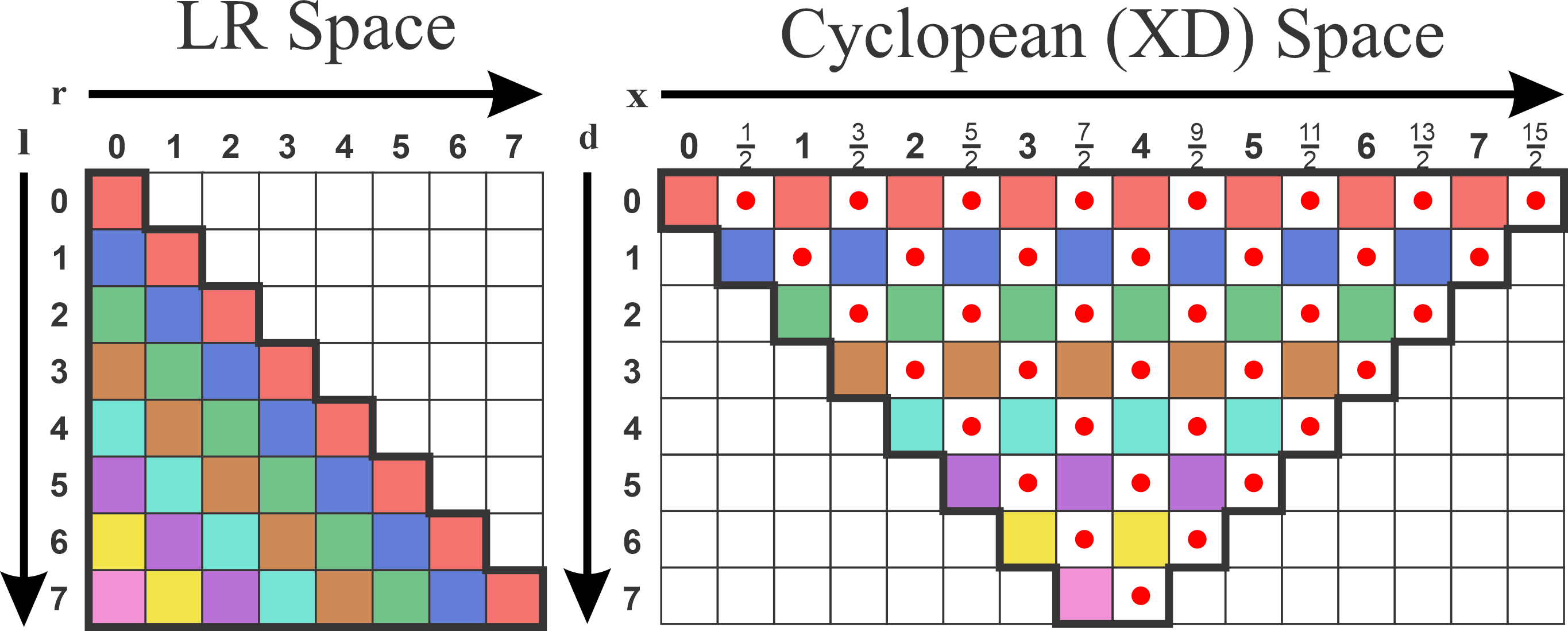}
   \caption{Space Transformation from L/R CS (left) to the XD (right). The colors represent a disparity value. Empty positions in XD space are disallowed, while the 'red dot'  data are obtained via a bilinear interpolation from LR data. The XD space has twice the resolution of the LR space, for each epipolar line.}
   \label{fig:space_transformation}
\end{figure}

\begin{figure}[ht]
    \centering
    \includegraphics[width=\columnwidth]{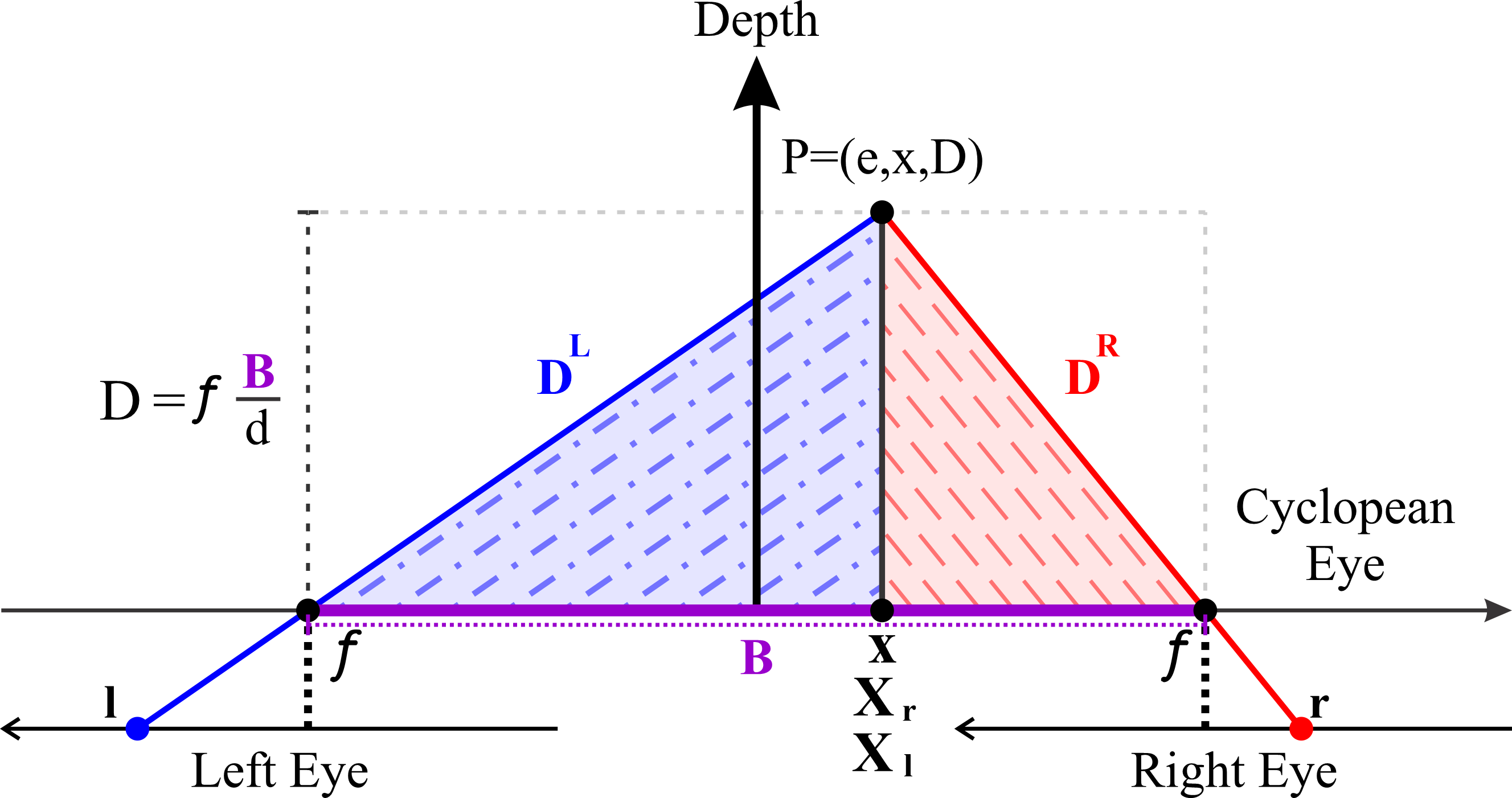}
    \caption{ $D^{L/R}(e,l/r)$ is the depth from L/R CS, respectively.  A point $P$ in 3D is described by the  XD as $P=P^{C}(e, x, Z={\cal D}(e,x ))$. The same point can be described by the L/R CS as $P=P^{L,R}(e,X_{l,r}, {\cal D}(e,x ))$, where $X_{l,r} \ne l,r$, since $l,r$ are the projective projection of $P$ into the L/R CS, while $X_{l,r}$ is the simpler orthogonal projection of $P$ into the L/R CS. Note that $B=|X_l-X_r|$. The distance to  $P$ measured by the L, R,  and cyclopean eye are ${\cal D}^L(e,l), {\cal D}^R(e,r), {\cal D}(e,x)$, respectively, and they are all different.}
    \label{fig:monocular_cyclopean_depth}
\end{figure}

Consider a left and right  images $I^L, I^R$ both with height and width $M \times  N$. A pixel coordinate system (CS) for the images are $\{ (e,l), (e,r) \}$  where $e \in (0,1,\hdots M-1)$ index their respective epipolar lines  and $l,r \in (0,1,\hdots N-1)$  (see Figure~\ref{fig:cyclopeaneye}). 
For many datasets, such as Middlebury~\cite{scharstein2014high}, the images have been rectified by the epipolar lines, which are then simply the horizontal lines of the images. The space $(e,x,d) \in M \times 2N \times D$ allows us to describe the matching of $(e,l) \leftrightarrow (e,r)$  as an assignment of a disparity $d$ to $(e,x)$ (see Figures~\ref{fig:space_transformation}, \ref{fig:monocular_cyclopean_depth}, and~\ref{fig:cyclopeaneye}). Such matching and associated assignment is described by the invertible coordinate transformation:
\begin{align}
    \begin{pmatrix}
     d   \\   x 
    \end{pmatrix} = \begin{pmatrix}
        1 & -1\\ \frac{1}{2} & \frac{1}{2} 
    \end{pmatrix} \begin{pmatrix}
        l \\ r
    \end{pmatrix}\, .
    \label{eq:cyclopean-transformation}
\end{align}

One consequence of this transformation is that the discrete cyclopean width coordinate  $x\in (0, \frac{1}{2}, 1,  \frac{3}{2}, \hdots,  N-1, N-\frac{1}{2})$ has subpixel resolution, twice as much as the image pixel width resolution of $l, r \in (0, 1, \hdots, N-1)$ (see Figure~\ref{fig:space_transformation}).

The XD provides a depth value ${\cal D}(e,x)$, for each match $(e,l) \leftrightarrow (e,r)$, as follows
\begin{align}
    {\cal D}(e,x) = f \frac{B}{d(e,x)}\, ,
    \label{eq:disparity-depth}
\end{align}
where $f$ is the focal length, $B$ is the baseline (distance between the left and right camera centers), see Figure~\ref{fig:monocular_cyclopean_depth}. 

\subsection{Depth from L, R, and XD}

From Figure~\ref{fig:monocular_cyclopean_depth} we infer the following triangle relations
\begin{align}
 {\cal D}^{L}(e,l)  =  \sqrt{ {\cal D}^2(e,x)  + \left(\frac{B}{2}-x\right )^2 }\quad \textrm{and} \nonumber \\
     {\cal D}^{R}(e,r)  =  \sqrt{{\cal D}^2(e,x)  + \left(\frac{B}{2}+x\right )^2 }
    \hspace{0.3in}
    \label{eq:depth-LR}
\end{align}
These relationships introduce a bias in the depth estimation from the L and R cameras relative to the XD depth values. To the best of our knowledge, these bias equations have not been derived previously. The impact becomes evident when comparing depth estimates with man-made GT scenarios. A consequence of mismatches in L/R depth estimation yield, for example, motion sickness in virtual reality~\cite{xia2024mismatch}.

\begin{proposition}[Datasets Depth in XD] 
\label{lemma:depth-data-sets}
The depth provided by the Middleburry dataset, and most man-made datasets, is  obtained by the match a light projector (or laser) offers followed by the formula \eqref{eq:disparity-depth}. Thus, it is not the depth from the L or R coordinate system.
\end{proposition}


\section{Image Features for Stereo Matching}
Image features are the first stage of processing for DL algorithms and in particular RAFT-Stereo~\cite{lipson2021raft} architecture allows other DL algorithms to extract their features, making them one of the most used features in deep learning.

A Bayesian interpretation for the use of features $F^L_{*}$ and $ F^R_{*}$ from left and right images, respectively,  is as follows
{\small 
\begin{align}
    P(I^L, I^R|S^{\textrm{3D}})&=\sum_{F^L, F^R}P(I^L, I^R,F^L, F^R|S^{\textrm{3D}}) 
    \nonumber \\
    & = \sum_{F^L, F^R}P(I^L, I^R|F^L, F^R) P(F^L, F^R|S^{\textrm{3D}})
     \nonumber \\
    & \approx P(I^L, I^R|F^L_{*}, F^R_{*}) P(F^L_{*}, F^R_{*}|S^{\textrm{3D}})
    \nonumber \\
    & = P(I^L, I^R|F^L_{*}, F^R_{*}) P(FM_{*}|S^{\textrm{3D}}) \, , \nonumber
\end{align}
}
where $F^L_{*}, F^R_{*}$ dominate the sum and RAFT-Stereo~\cite{lipson2021raft} also does the following computations
{\small
\begin{align}
FM(e,x,d(e,x))=1 - \frac{FMS(e,x,d(e,x))}{max(FMS(e,x,d(e,x))},\quad \textrm{where}
 \nonumber \\
   FMS(e,x,d(e,x))=F^L(e,x-\frac{1}{2}d(e,x))\cdot F^R(e,x+\frac{1}{2}d(e,x))\, , \nonumber
\end{align}
}
and $d(e,x)$ is the cyclopean disparity that describes the 3D surface. Thus, the feature matching similarity (FMS) must be large when there is a good feature match and $FM(e,x,d(e,x))\in [0,1]$ is used so that good matches render these values small and bounded to the $[0,1]$ range. 

In the Bayesian view, state-of-the-art (SOTA) DL algorithms, e.g., RAFT-Stereo~\cite{lipson2021raft}, break the stereo problem $\arg \max_{S^{3D}} P(S^{\textrm{3D}}|I^L, I^R) $ in two stages, first extracting features $F^L_{*}, F^R_{*}$ from left and right images and then performing in the second stage
{\small
\begin{align}
\arg \max_{S^{3D}} \left [ P(S^{\textrm{3D}}|FM_{*})\right ]=\arg \max_{S^{3D}} \left [ P(FM_{*}|S^{\textrm{3D}}) \, P(S^{\textrm3D})\right ]
    \label{eq:bayesian-stereo-feature}
\end{align}
}
Figures~\ref{fig:occlusions},~\ref{fig:uniqueness},~\ref{fig:homogeneity}, and~\ref{fig:repetitive_patterns} show the quality of their extracted $FM_{*}(e,x,d(e,x))$  yielding a fairly sparse description in the matching space. In various regions, where data matching is available, it allows a visualization of the $S^{\textrm{3D}}$ solution that maximizes the likelihood $P(FM_{*}|S^{\textrm{3D}})$. 

Our data investigations led us to Figure~\ref{fig:repetitive_patterns} where multiple matches occur associated with a left image feature or with a right image feature. However, a final solution must choose only one of them to be correct. In Figure~\ref {fig:uniqueness} multiple matches occur for the same Cyclopean Eye coordinate $x$ and in this case, human perception chooses one match. Also, there are no good feature matches in regions of occlusions such as in Figure~\ref{fig:occlusions} and neither on homogeneous regions such as in Figure~\ref{fig:homogeneity}. Thus, all these scenarios lead us to conclude that for a $S^{\textrm 3D}$ solution, given the features,  there is a  need for a prior $P(S^{\textrm3D})$ to help solve for the stereo problem, which we study next. 
\section{Geometrical Constraints (GCs)}

We will now address three geometrical constraints that provide the basis for our understanding and analysis of geometrical properties in stereo vision.  When creating a stereo algorithm these constraints could be expressed by a prior $P(S^{3D})$ present in equation~\eqref{eq:bayesian-stereo-feature}. However, in this paper we do not attempt to create a stereo algorithm.  The format we present this section is by introducing the definitions needed for us to articulate a geometrical constraint proposition followed by an analysis with real datasets. 

\begin{definition}[Opaque Surfaces and Stereo] 
\label{def:opaque-surfaces}
3D opaque surfaces do not let light pass through them. Consider a 3D point $P=(X,Y,Z)$ that belongs to an opaque surface. If it is visible by L or R, we can describe it as $(e,l, {\cal D}^L(e,l))$ or as $(e,r, {\cal D}^R(e,r))$, respectively. 
\end{definition}
Note that if $P$ is visible by both eyes, as illustrated in Figure~\ref{fig:monocular_cyclopean_depth}, we have $(e,l) \leftrightarrow (e,r=l-d^L(e,l))$ and so (i) the disparities satisfy $d^L(e,l)=d^R(e,r=l-d^L(e,l))$, but in general ${\cal D}^L(e,l)\ne {\cal D}^R(e,r)$; and (ii) the assignment $d(e,x)$ derived from \eqref{eq:cyclopean-transformation} yields ${\cal D}(e,x)$ from \eqref{eq:disparity-depth}, but ${\cal D}(e,x)\ne {\cal D}^L(e,l),  {\cal D}^R(e,r)$ as seen from \eqref{eq:depth-LR}.

\begin{definition}[Transparent Surfaces and Stereo] 
\label{def:transparent-surfaces}
Transparent surfaces allow some light to pass through. 
Two distinct points $P_{1,2}=(X_{1,2},Y_{1,2},Z_{1,2})$ are in a transparent pair state if both can be seen by the XD and share the same $(e,x)$ coordinates. Transparent stereo surfaces are made of a set of contiguous transparent-pair states. 
\end{definition}

\begin{figure}[!ht]
    \centering
    \includegraphics[width=\columnwidth]{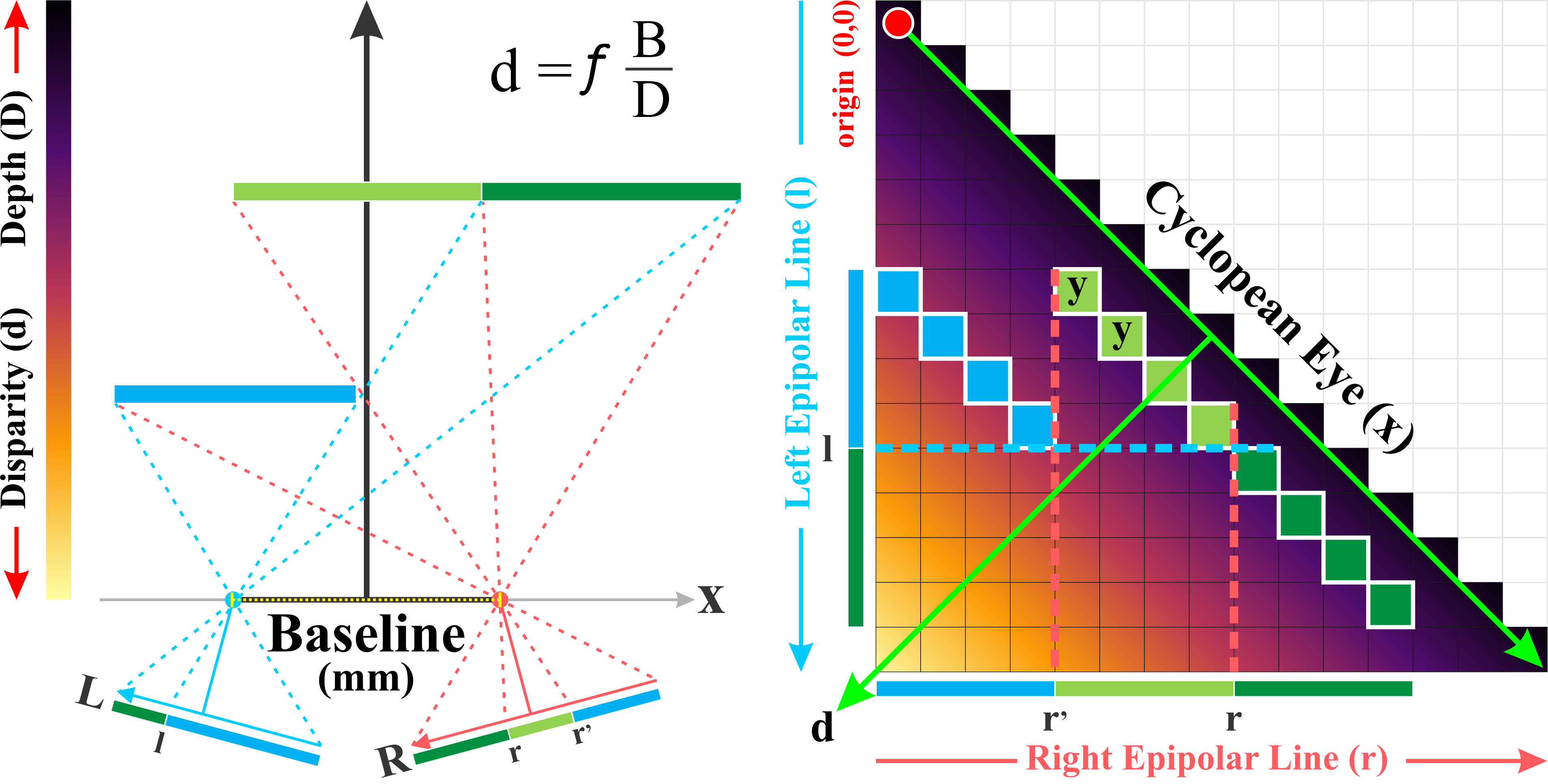}
    \caption{An epipolar slice of a surface with left occlusion region and its description by the XD. Left. A top view of the epipolar slice of the surface and the two eyes projections. The baseline ${\bf B}$  connects L to R focal centers. The depth axis describe the inverse of disparity (Equation \eqref{eq:disparity-depth} depicted). Right. A discrete XD, a rotation of the L/R CS described by Equation~\eqref{eq:cyclopean-transformation}. Note that the L CS is pointing down. The two red vertical dashed lines delimit the R occlusion area which are associated with a L discontinuity along the horizontal blue dashed line with a jump of the same size as the right occlusion, as described by Da Vinci-GC in Proposition~\ref{prop:Da_Vinci-GC}. Note that the only two (2) light green squares (the ones without an "y" in them) are seen by the XD associated with the R occlusion, satisfying one disparity per coordinate $x$ (as postulated by Opaque-GC), which is {\bf half} of the size of the R occlusion area. }
    \label{fig:cyclopeaneye}
\end{figure}

\begin{figure*}[!ht]
    \centering
    \includegraphics[width=\textwidth]{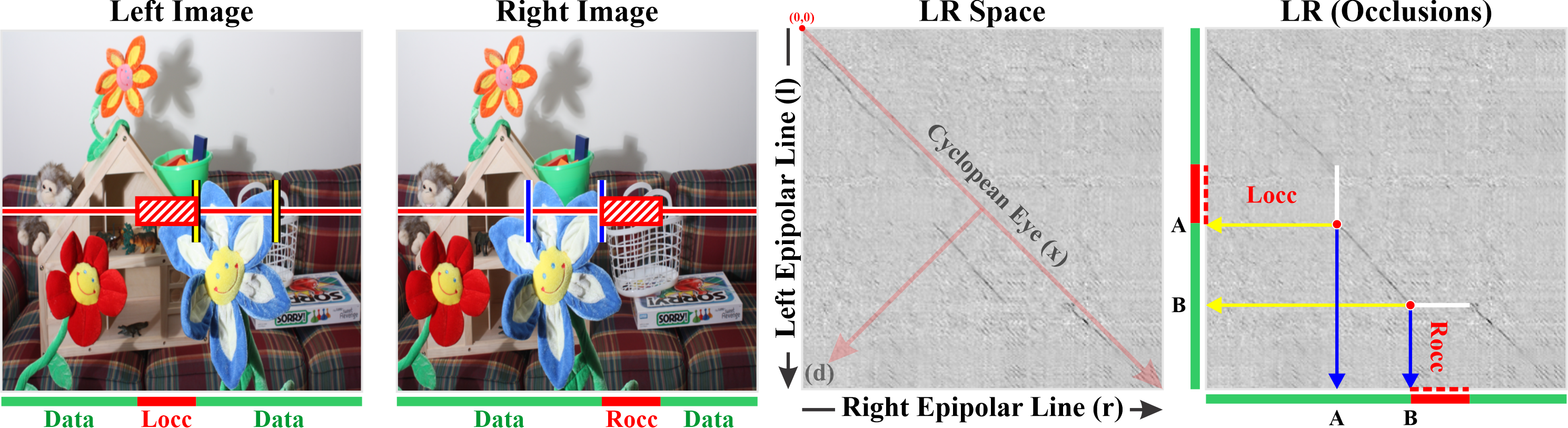}
    \caption{\textbf{Occlusions:} The LR space displays $FM$ distances associated with the epipolar lines $e=128$ (red), where $FM_{e,x}(d) \in [0,1]$. Dark regions (low values of $FM$) represent good matches. Da Vinci-GC is verified with L/R-occlusions in dashed-red, associated with the L/R-discontinuities in white (jumps). In the cyclopean eye these are 45 degrees jumps. Moreover, the Opaque-GC (unique disparity per cyclopean eye coordinate) is also verified. This figure also presents an association of the LR Space data/occlusions with the RGB LR images, where the points A and B in the LR Space denotes the start/end of the blue flower.}
    \label{fig:occlusions}
\end{figure*}

With these two definitions, we state a geometrical constraint
\begin{proposition}[Opaque-GC]
\label{prop:Opaque-GC}
 For opaque surfaces each cyclopean coordinate $(e,x)$ has one and only one disparity, i.e.,  $d$ is a function $d:(e,x) \rightarrow {\mathbb R}$.
\end{proposition}
Note that this constraint is to be interpreted as prior geometrical constraint. Figure~\ref{fig:cyclopeaneye} illustrates this constraint, while Figure~\ref{fig:occlusions} shows data that validate this constraint. 


\begin{figure*}[!ht]
    \centering
    \includegraphics[width=\textwidth]{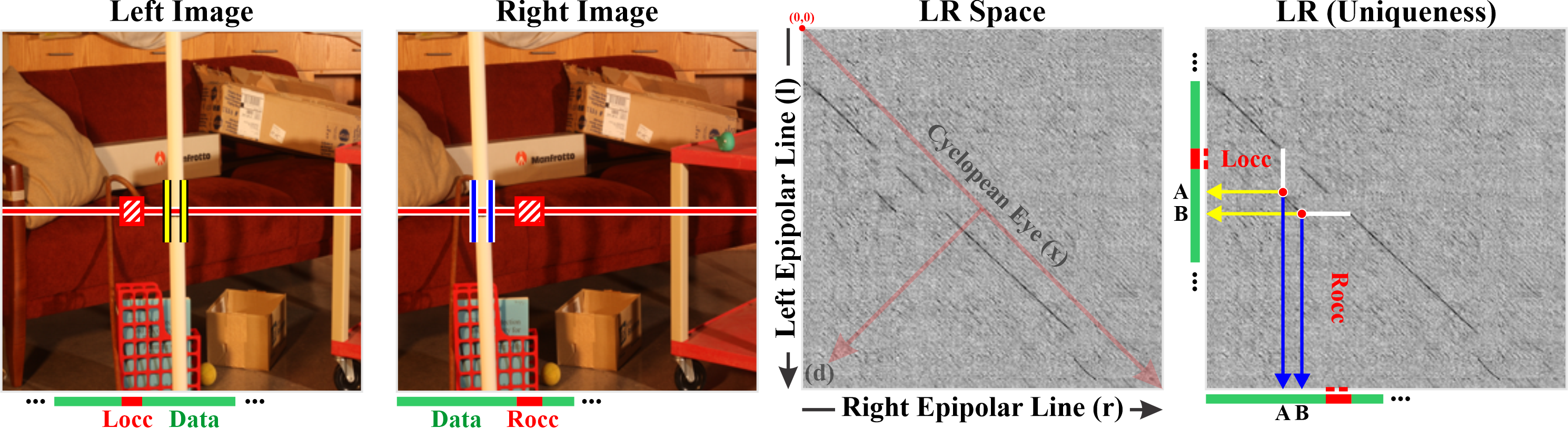}
    \caption{\textbf{Uniqueness:} The LR Space displays $FM$ distances associated with the epipolar lines $e=128$ (red), where $FM_{e,x}(d) \in [0,1]$. The Opaque-GC requires that either the solution recovers the thin pole in front or ignore it. On the right we show the solution that recover the thin pole in front with Da Vinci-GC, i.e., with the L/R-discontinuities in white (jumps). From the cyclopean eye these are 45 degrees changes. The Opaque-GC is consistently applied. Note that L/R-occlusions in dashed-red end up being expanded to regions where other possible good feature matches exist (low $FM$ values). This figure also presents an association of the LR Space data/occlusions with the RGB LR images, where the points A and B denotes the start/end of the pole.}
    \label{fig:uniqueness}
\end{figure*}

However, a scenario with opaque surfaces  may not satisfy this rule. More precisely, it is possible in some special scenarios for two 3D points $P_1,P_2$, visible by the left eye as $(e,l_1), (e,l_2)$ and by the right eye as $(e,r_1), (e,r_2)$, to satisfy that both have the same $x=\frac{l_1+r_1}{2}=\frac{l_2+r_2}{2}$ but different disparities, for example, for the case of thin objects in front of backgrounds, as shown in Figure~\ref{fig:uniqueness}. Thus, it is suggested that this rule is applied to recover surfaces. In these scenarios the above constraint, used as prior knowledge, will then force the solution to choose either one pair of matches or the other, but not both.

\begin{figure*}[!ht]
   \centering
   \includegraphics[width=\textwidth]{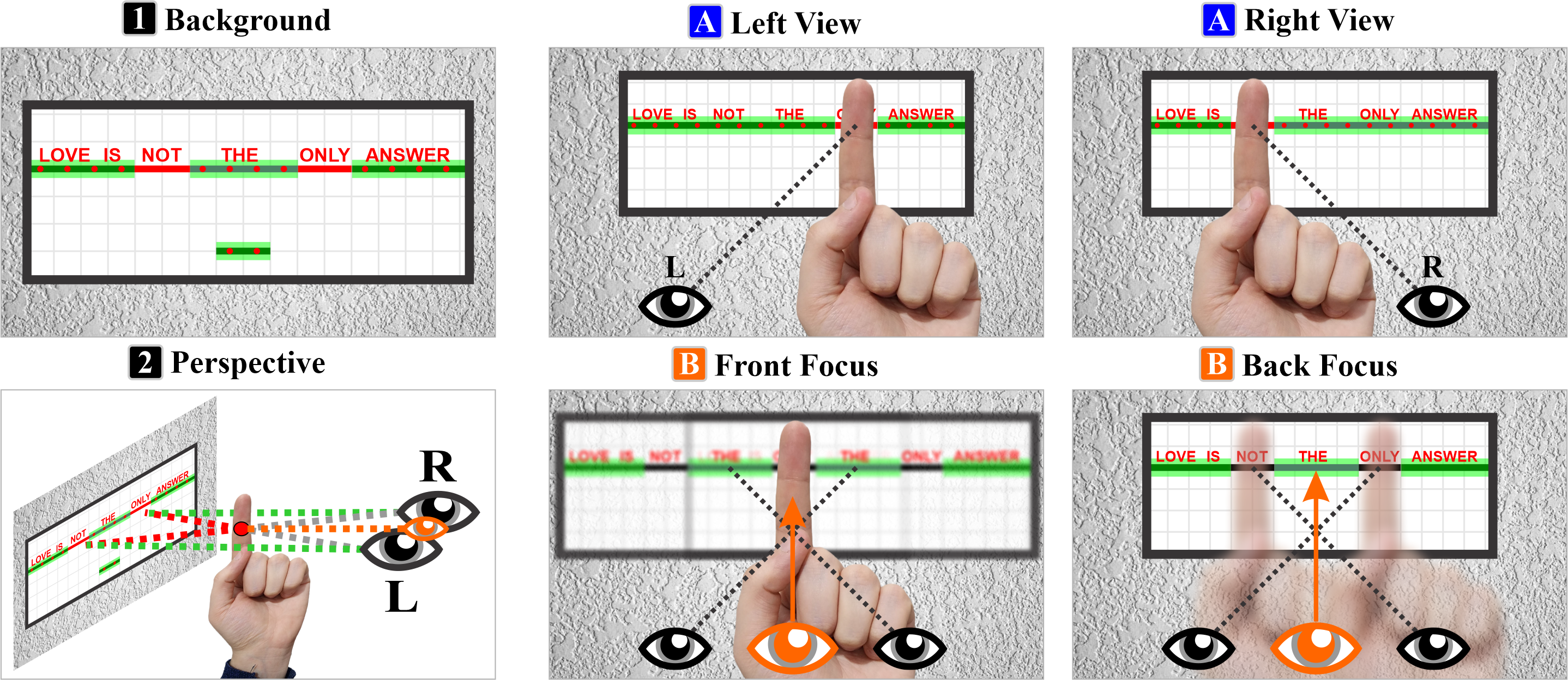}
   \caption{\textbf{Human perception experiment:} This figure illustrates a stereo vision experiment with two objects in a scene: a left hand and finger (front) and a text frame on a wall (back). 1) The top-left subfigure shows the wall, and the text frame containing the phrase "Love is not the only answer". The green regions (e.g the sequence of words 'Love', 'is', 'the', and 'answer') represent the text seen by both L/R views, while the red regions (e.g the words 'not', and 'only') represent half occlusions. 2) The bottom-left subfigure shows the experiment in another perspective, where the red intersection point represents the position of the front object (finger) exactly in the middle of L/R eyes, that is, facing the Cyclopean Eye (orange). The finger cause an occlusion into the right and left eyes in the words 'not' and 'only', respectively. When analyzing the left and right views individually (subfigures Blue-A)  the left eye sees the phrase 'Love is not the answer', the right eye sees the phrase 'Love is the only answer'. The  human cyclopean eye (subfigures Orange-B) decides which object to pay attention, either directing the focus into the finger (front) or in the text frame (back). If the attention (Front Focus) is at the finger (front), the background seems to be blurred and duplicated. However, if the attention (Back Focus) is at the Background (wall + text frame), the finger (front) seems to be transparent, blurred and duplicated. Interestingly, by the cyclopean view it is possible to combine the data from the L/R view, then it is possible to see the entire phrase 'Love is not the only answer'. Observe that this experiment is in accordance with the propositions 2 and 3. Also, the choice of focusing on the finger data offers a solution similar to the scenario in Figure~\ref{fig:uniqueness}.}
   \label{fig:perception}
\end{figure*}

Human perception faces this "dilemma" when, for example, placing a finger in front of a scene (see Figure~\ref{fig:perception}). In this case, humans can either focus on the finger or in the background and reconstruct the 3D structure of the finger (ignoring background) or background (ignoring the finger). It is the attention mechanism that makes the choice of how to apply the GC-constraint and ignore the alternative solution, and today we do not see any attention mechanism being used by SOTA stereo.

\begin{definition}[Occlusions and Discontinuities]
R- (L-) occlusions are regions that are seen by the R (L) eye but not by  L (R).  R- (L-) discontinuities are places where jumps of disparity (or of depth) occur in the R (L).  L- and R-occlusions are termed half-occlusions~\cite{Belhumeur96, Wang_2019_CVPR}. 
\end{definition}

As first observed by Da Vinci 3D opaque surfaces follow GCs that link discontinuities to occlusions, see also~\cite{Geigeretal92, Geigeretal95, Belhumeur96, IshikawaGeiger98}. 
We next describe our new proposed GCs. 
\begin{proposition}[Da Vinci-GC]
\label{prop:Da_Vinci-GC}
 In the cyclopean eye the jumps along an epipolar line must occur along the $45$ degree angles.
 Equivalently, the size of the jump of a R- (L-) discontinuity is equal to the size of the L- (R-) occlusion. 
\end{proposition}

Figure~\ref{fig:cyclopeaneye} illustrates this constraint.  
Figure~\ref{fig:occlusions} and Figure~\ref{fig:uniqueness} show and provide an analysis of the disparity jumps of 45 degrees. Da Vinci-GC was proposed in the works \cite{Geigeretal92,IshikawaGeiger98}. To the best of our knowledge, Da Vinci-GC in the cyclopean eye has not been proposed before. Figure~\ref{fig:uniqueness}  illustrates the consistency between Da Vinci-GC and Opaque-GC, where a solution that satisfies one constraint also satisfies the other constraint, both constraints forcing a solution to choose which data to focus on and which data to ignore. 

Note that if a prior with this constraint is to be applied at some stage of a stereo algorithm, then the final disparity values assigned at such occlusion locations must be processed at a later stage. Feature matching also fails at homogeneous regions as we discuss next.

\begin{definition}[Homogeneous Regions]
Homogeneous regions are areas in an image that lack texture. All stereo feature responses in these regions provide equally poor matches.
\end{definition}

\begin{figure*}[!ht]
    \centering
    \includegraphics[width=\textwidth]{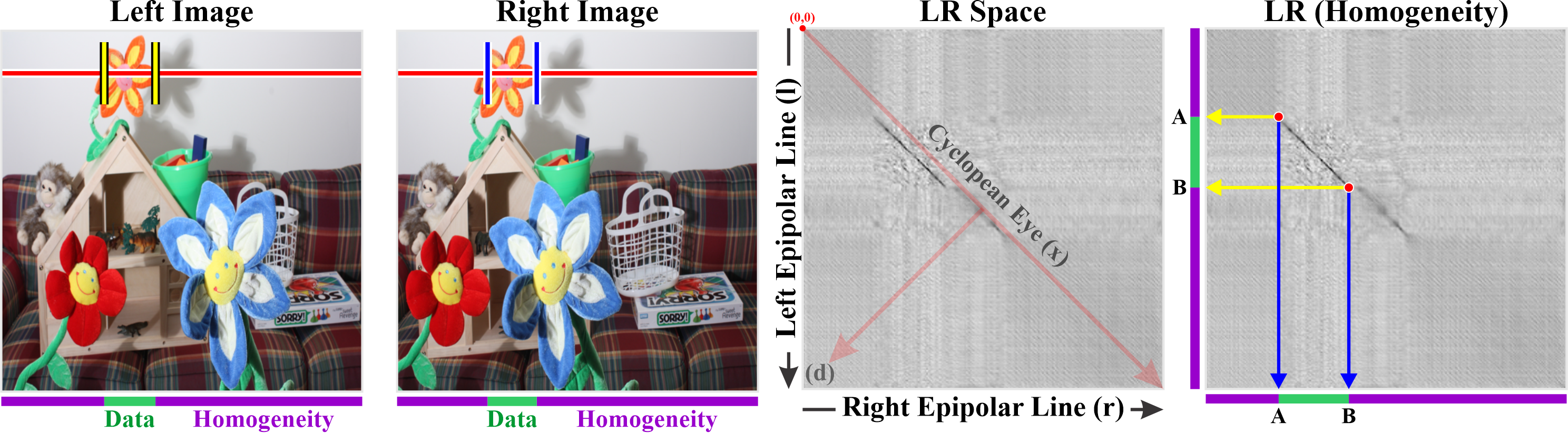}
    \caption{\textbf{Homogeneity:} The LR space displays $FM$ distances associated with the epipolar lines $e=30$ (red), where $FM_{e,x}(d) \in [0,1]$. Dark regions represent low $FM$ and good matches. Note that shadows do yield good matches. However, in order to fill in homogeneous regions, surface priors become necessary.}
    \label{fig:homogeneity}
\end{figure*}

Figure~\ref{fig:homogeneity} illustrates homogeneous regions in real data. Feature matching at homogeneous regions does not discriminate which disparity to choose. Thus, feature matching does not provide data do discriminate the disparity for homogeneous regions nor for occlusion regions. 

\begin{definition}[Repetitive Patterns]
Objects with similar textures may create repeated patterns (multiple matching possibilities) in the LR and XD space. Although stereo feature responses in these regions may provide equally good matches, only the correspondence that correctly aligns the beginning and end of the object represents the true solution.
\end{definition}

\begin{figure*}[!ht]
    \centering
    \includegraphics[width=\textwidth]{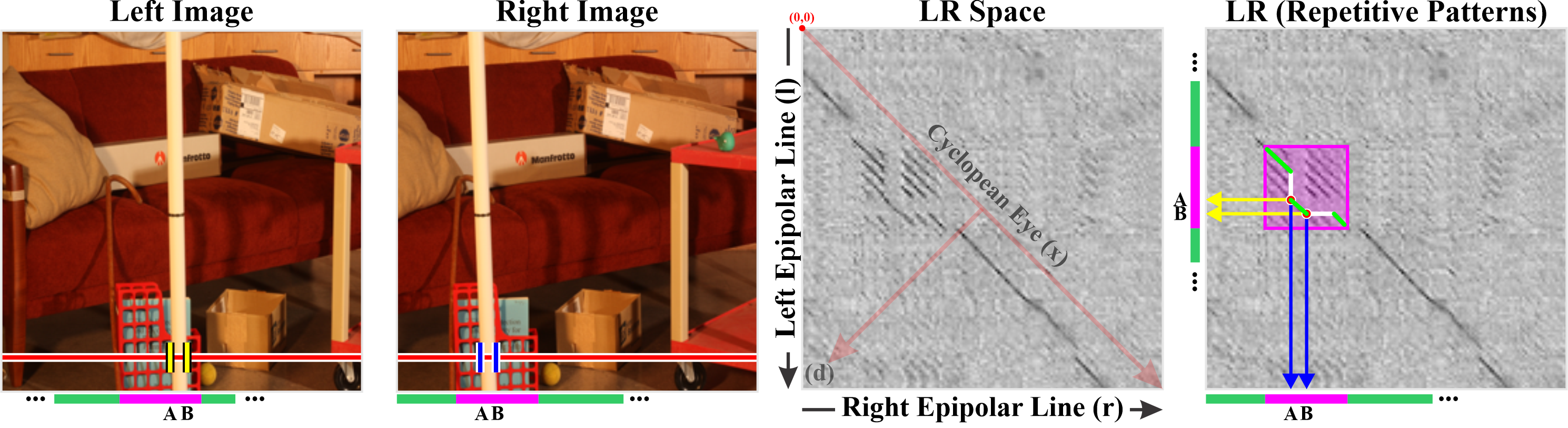}
    \caption{\textbf{Repetitive Patterns:} The LR space displays $FM$ distances associated with the epipolar lines $e=464$ (red), where $FM_{e,x}(d) \in [0,1]$. Dark regions represent low $FM$ and good matches. Note the red object behind the pole with a repetitive pattern in the LR space, causing multiple good matches. However, an optimal solution which align with the beginning (left-most part) of the object and satisfying the geometrical constraints is the ground truth solution represented by light-green inside the pink box in the right-most subfigure.}
    \label{fig:repetitive_patterns}
\end{figure*}

Figure~\ref{fig:repetitive_patterns} illustrates repetitive patterns caused by an object composed of similar visual structures and textures. This example demonstrates a scenario where feature matching alone is insufficient to determine the correct disparity. Unlike homogeneous regions, which lack strong matching candidates, repetitive pattern regions yield multiple false-positive matches.

\begin{proposition}[Filling-GC]
\label{prop:Filling-Constraint-3}
For regions where the data matching does not offer a clear solution, a prior geometrical constraint satisfying Da Vinci-GC should be applied to recover the disparity function $d(e,x)$ everywhere in the cyclopean eye. 
\end{proposition}

We conclude this section by pointing out that our submitted paper [Under Review] did an analysis of Filling-GC for GT data and DL algorithms. Our study demonstrated that the best techniques nowadays on the main benchmarks tend to reconstruct surfaces with low Gaussian Curvature as a prior. The new generation of stereo approaches leverage stereo matching, and also provide surface consistency by combining monocular features (e.g from DepthAnythingV2~\cite{yang2024depthv2}).


\section{Conclusion}
\label{sec:conclusion}
We presented an analysis of the geometric foundations underlying 3D stereo vision. Starting with the Cyclopean Eye model, we incorporated geometric constraints that account for depth discontinuities and occlusions, showing how this model combines information from both left and right eyes into one unified perspective. The Cyclopean Eye, in turn, supports the attention-based mechanisms employed by human stereo vision.

Our investigation emphasized the role of geometric structure in understanding and interpreting 3D scene reconstruction. We analyzed the quality of stereo feature matching derived from deep learning models, proposed novel geometric constraints tailored to stereo vision, and highlighted how these constraints help interpret the performance of data driven methods.

Together, these contributions reinforce the idea that combining strong geometric priors with learned features leads to more interpretable, generalizable, and reliable stereo vision systems. Ultimately, our goal is to underscore the importance of 3D geometric modeling in capturing critical visual information and guiding the development of next-generation vision approaches.

\section*{Acknowledgment}
This study was financed in part by the Coordenação de Aperfeiçoamento de Pessoal de Nível Superior - Brasil (CAPES) - Finance Code 001. 



\bibliographystyle{IEEEtran}
\bibliography{main}
%
%


\end{document}